%% file: main.tex
\documentclass[letterpaper]{article} 
\usepackage{aaai2027}  
\usepackage[hyphens]{url}  
\usepackage{graphicx} 
\usepackage{natbib}  
\usepackage{caption} 
\usepackage{algorithm}
\usepackage{algorithmic}
\usepackage{amsmath} 
\usepackage{newfloat}
\usepackage{listings}
\DeclareCaptionStyle{ruled}{labelfont=normalfont,labelsep=colon,strut=off} 
\floatstyle{ruled}
\newfloat{listing}{tb}{lst}{}
\floatname{listing}{Listing}

\usepackage{booktabs}

\nocopyright

\title{CopyCat: Improving Fine-Grained Subject Consistency in Subject-to-Image Models within Seconds}
\author{
    Peng Zheng$^{1,2}$\quad\quad
    Ruiqi Liu$^{1}$\quad\quad
    Rui Ma$^{1,4,}$\thanks{Corresponding authors}\quad\quad
    Zuxuan Wu$^{2,3,}$\footnotemark[1]
}
\affiliations{
    \textsuperscript{\rm 1}School of Artificial Intelligence, Jilin University\quad\quad
    \textsuperscript{\rm 2}Shanghai Innovation Institute\\
    \textsuperscript{\rm 3}Institute of Trustworthy Embodied AI, Fudan University\\
    \textsuperscript{\rm 4}Engineering Research Center of Knowledge-Driven Human-Machine Intelligence, MOE, China\\
}

\begin{document}

\maketitle

\input{sec/0_abs}
\input{sec/1_intro}
\input{sec/2_related}

\input{sec/3_method}

\input{sec/4_exp}
\input{sec/5_conclusion}

\bibliography{main}

\end{document}

%% file: sec/0_abs.tex
\begin{abstract}
Recent subject-to-image models have achieved impressive progress in personalized image generation, yet they still struggle to preserve fine-grained subject-specific details. A major reason is the lack of high-quality fine-grained identity supervision: real paired data are expensive to collect, while synthesized training pairs often preserve only coarse subject appearance and fail to capture subtle subject-specific details. In this work, we propose CopyCat, a lightweight model-refinement framework that improves fine-grained subject consistency within only a few seconds. CopyCat performs a one-time refinement of a pretrained subject-to-image model by attaching a lightweight Fine-grained Consistency LoRA (FCLoRA) and optimizing it using a single proxy image, which is used as both the conditioning image and the reconstruction target. This exact self-reconstruction objective substantially simplifies the optimization task, enabling effective fine-grained refinement within only a few seconds. The refinement is performed only once; the resulting model can be directly applied to diverse unseen reference subjects and prompts without further subject-specific optimization. We further revisit subject-to-image LoRA training in double-stream diffusion transformers and find that adapting only the visual stream consistently improves subject consistency. Extensive experiments on DreamBench and XVerseBench demonstrate consistent improvements in fine-grained subject consistency across representative subject-to-image models under both single- and multi-subject settings.
\end{abstract}

%% file: sec/1_intro.tex
\begin{figure*}
    \centering
    \includegraphics[width=\linewidth]{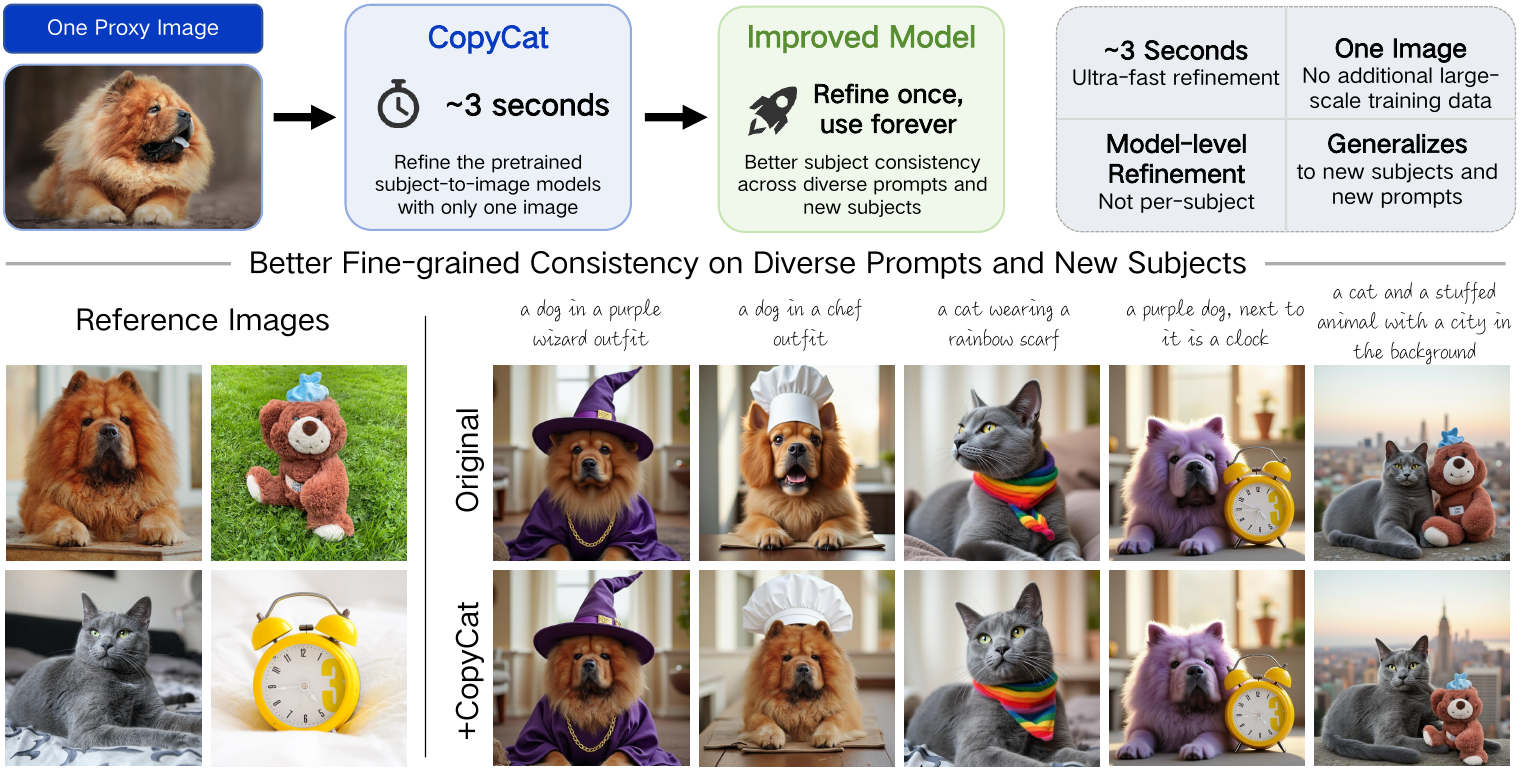}
    \caption{\textbf{Overview of CopyCat.}
Using a single proxy image, CopyCat learns a lightweight
Fine-grained Consistency LoRA to perform a one-time,
few-second refinement of a pretrained subject-to-image model.
The refined model is directly applicable to unseen reference
subjects and diverse prompts without further subject-specific
optimization. Representative examples demonstrate improved
preservation of fine-grained subject details while maintaining
prompt fidelity.}
    \label{fig:teaser}
\end{figure*}

\section{Introduction}

Recent advances in text-to-image generation have enabled remarkable progress in high-quality visual content synthesis. Building upon these powerful generative models, subject-driven image generation aims to synthesize novel images that not only follow textual instructions but also faithfully preserve the identity of one or more reference subjects, making it valuable for applications such as personalized content creation, virtual photography, and digital entertainment. Early personalization approaches, including DreamBooth~\cite{ruiz2023dreambooth} and Textual Inversion~\cite{gal2022image}, optimize the model for each target subject, while more recent subject-to-image (S2I) models, such as UNO~\cite{wu2025less}, enable feed-forward generation by learning subject-conditioned generation from large-scale training data. Despite these advances, existing subject-to-image models still struggle to faithfully preserve fine-grained subject-specific details, limiting the fidelity of personalized image generation.

A major obstacle lies in the supervision itself. Existing subject-to-image models are primarily trained using large-scale synthesized subject-image pairs because collecting real paired data with accurate identity correspondence is prohibitively expensive. Although synthesized data provide effective supervision for coarse subject appearance, they often fail to preserve subtle subject-specific details. Consequently, current models often preserve the coarse identity of the reference subject while failing to faithfully reproduce the fine-grained details that distinguish one subject from another.

One straightforward solution is to improve the quality and scale of synthesized training data. However, this requires repeatedly constructing large-scale datasets and retraining expensive subject-to-image models, making it difficult to quickly benefit existing pretrained models. Instead, we ask a different question: \emph{can an existing pretrained subject-to-image model recover its missing fine-grained subject consistency through an extremely lightweight refinement without requiring additional large-scale training data?}

Motivated by this question, we propose \textbf{CopyCat}, a lightweight model-refinement framework that significantly improves fine-grained subject consistency within only a few seconds. As illustrated in Figure~\ref{fig:teaser}, CopyCat performs a one-time refinement of an existing subject-to-image model by introducing a lightweight \emph{Fine-grained Consistency LoRA} (FCLoRA). During refinement, a single image, referred to as a \emph{proxy image}, is used as both the conditioning image and the reconstruction target, forming an exact self-reconstruction objective. Surprisingly, we find that this simple objective is effective only when the conditioning image and the reconstruction target are exactly identical; any mismatch, such as using another image of the same subject or changing the image resolution, causes the refinement to fail to produce noticeable improvement. Because the conditioning image and the reconstruction target are perfectly aligned, the refinement objective eliminates the need to model subject variations, allowing the lightweight LoRA to focus exclusively on strengthening fine-grained visual correspondence. Rather than learning new subject knowledge, CopyCat recovers the underutilized fine-grained consistency capability already embedded in the pretrained subject-to-image model. This refinement is performed only once. The resulting model can subsequently be applied to diverse unseen reference subjects and prompts without subject-specific optimization.

Beyond the proposed model refinement, we further revisit the design of subject-to-image LoRA in existing double-stream diffusion transformers. We find that adapting the text stream is unnecessary for learning subject-specific visual characteristics. Instead, removing LoRA from the dedicated text stream while retaining it in the image stream consistently improves subject consistency while reducing the number of trainable parameters. This simple design provides a stronger subject-to-image LoRA and serves as an independent improvement that can be readily adopted during subject-to-image model training.

Our contributions are summarized as follows:

\begin{itemize}
    \item We propose \textbf{CopyCat}, a lightweight model-refinement framework that substantially improves fine-grained subject consistency through only a few seconds of one-time refinement.

    \item We identify exact self-reconstruction using a single proxy image as the key to effective few-second model refinement. This finding provides a simple yet effective paradigm for improving fine-grained subject consistency without requiring additional large-scale training data.

    \item We revisit the LoRA design in double-stream subject-to-image models and demonstrate that applying LoRA only to the image stream outperforms the conventional strategy that adapts both the visual and text streams.
\end{itemize}

%% file: sec/2_related.tex
\section{Related Work}

\subsection{Optimization-based Subject Personalization}

Early subject personalization methods customize a pretrained text-to-image model through subject-specific optimization. Textual Inversion~\cite{gal2022image} learns customized textual embeddings while keeping the diffusion model frozen, whereas DreamBooth~\cite{ruiz2023dreambooth} fine-tunes the model using only a few reference images with prior preservation. HyperDreamBooth~\cite{ruiz2024hyperdreambooth}, Custom Diffusion~\cite{kumari2023multi}, Perfusion~\cite{tewel2023key}, and LoRA-based adaptation~\cite{hu2022lora} further improve personalization efficiency by reducing the optimization cost or the number of trainable parameters. To support multi-concept personalization, subsequent studies explore attention manipulation, concept decomposition, and parameter composition, including Break-A-Scene~\cite{avrahami2023break}, Cones~\cite{liu2023cones,NEURIPS2023_b3847cda}, Mix-of-Show~\cite{gu2023mix}, Concept Weaver~\cite{kwon2024concept}, ZipLoRA~\cite{shah2024ziplora}, Orthogonal Adaptation~\cite{po2024orthogonal}, CLoRA~\cite{meral2025contrastive}, ConceptSplit~\cite{lim2025conceptsplit}, and FreeLoRA~\cite{zheng2025freelora}. Although these methods achieve high subject fidelity, they require subject-specific optimization for every new subject, resulting in considerable optimization cost and limited scalability.

\begin{figure*}[!t]
    \centering
    \includegraphics[width=\textwidth]{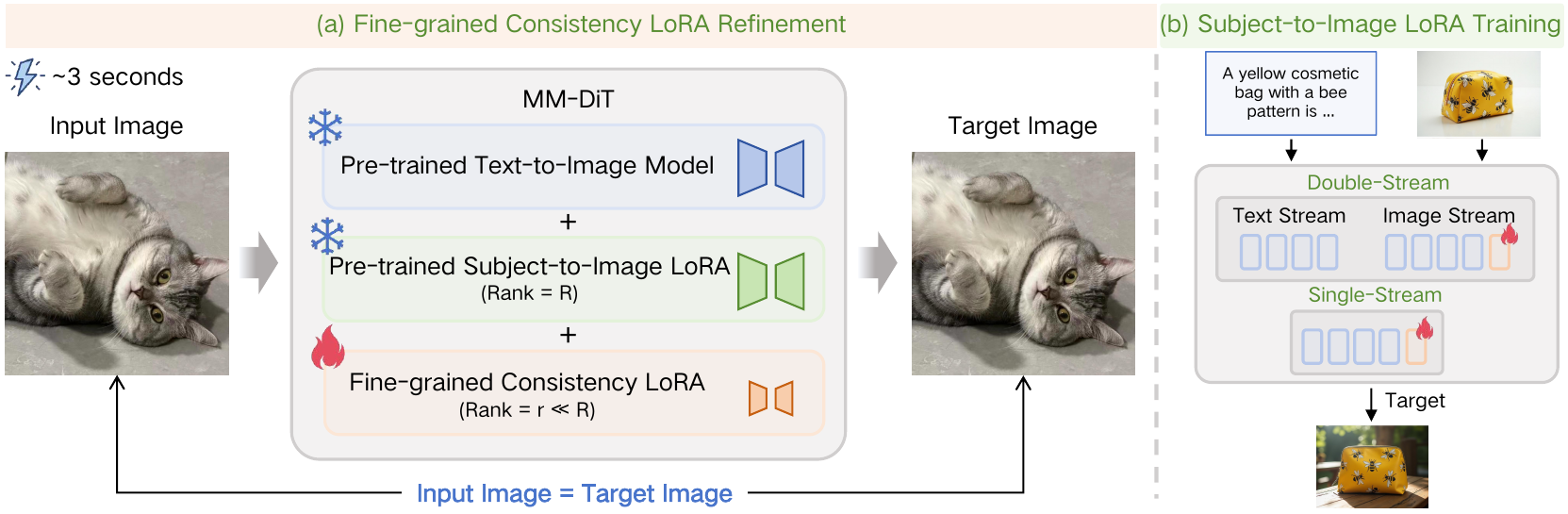}
    \caption{
    \textbf{Overview of the proposed method.}
    (a) CopyCat attaches a lightweight Fine-grained Consistency LoRA (FCLoRA) to a frozen subject-to-image model and performs a one-time refinement completed within only a few seconds using a single proxy image, where the conditioning image and the reconstruction target are pixel-wise identical.
    (b) We revisit the design of Subject-to-Image LoRA in double-stream diffusion transformers and remove LoRA from the dedicated text stream while retaining it in the image stream and the subsequent shared Single-Stream Blocks.
    }
    \label{fig:pipeline}
\end{figure*}

\subsection{Feed-forward Subject-to-Image Generation}

Feed-forward methods eliminate subject-specific optimization by directly extracting subject representations from one or more reference images. Representative approaches include ELITE~\cite{wei2023elite}, BLIP-Diffusion~\cite{li2023blip}, SuTI~\cite{chen2023subject}, IP-Adapter~\cite{ye2023ip}, PhotoVerse~\cite{chen2023photoverse}, Taming Encoder~\cite{jia2023taming}, PhotoMaker~\cite{li2024photomaker}, InstantBooth~\cite{shi2024instantbooth}, InstantID~\cite{wang2024instantid}, and PuLID~\cite{guo2024pulid}. To support multiple customized subjects, Subject Diffusion~\cite{ma2024subject}, FastComposer~\cite{xiao2025fastcomposer}, DreamMatcher~\cite{nam2024dreammatcher}, MS-Diffusion~\cite{wang2025ms}, MIP-Adapter~\cite{huang2025resolving}, AnyStory~\cite{he2025anystory}, and MoA~\cite{wang2024moa} improve subject disentanglement through subject-aware conditioning, regional attention, or unified modulation. More recently, diffusion-transformer-based frameworks, including OminiControl~\cite{tan2025ominicontrol}, OmniGen~\cite{xiao2025omnigen}, OmniGen2~\cite{wu2025omnigen2}, DreamO~\cite{mou2025dreamo}, UNO~\cite{wu2025less}, XVerse~\cite{chen2026xverse}, and USO~\cite{wu2025uso} further improve image customization through unified multimodal modeling. These approaches eliminate subject-specific optimization but remain highly dependent on large-scale subject-image training data, where fine-grained identity supervision is often limited.

\subsection{Post-training Subject Consistency Refinement}

Recently, post-training refinement has emerged as a promising paradigm for further improving existing subject-to-image models. FocusDPO~\cite{jin2026focusdpo} introduces dynamic preference optimization to enhance subject fidelity and suppress attribute leakage in pretrained personalized generation models. PSR~\cite{Wang_2026_CVPR} further improves multi-subject personalization by introducing pairwise subject-consistency rewards together with reinforcement learning for post-training optimization. UMO~\cite{cheng2026scaling} scales post-training optimization through a multi-to-multi matching paradigm with large-scale synthesized and real preference data. USO~\cite{wu2025uso} extends this direction by incorporating Style Reward Learning into a unified framework for subject- and style-driven generation.

Different from these approaches, which rely on large-scale preference optimization or reward learning, CopyCat performs a one-time refinement completed within only a few seconds using an exact self-reconstruction objective constructed from a single proxy image. By introducing a lightweight Fine-grained Consistency LoRA, our method effectively improves fine-grained subject consistency without requiring additional large-scale training data. The resulting refined model can be directly applied to unseen reference subjects without further subject-specific optimization.

%% file: sec/3_method.tex
\section{Method}

Figure~\ref{fig:pipeline} illustrates the proposed framework, which consists of two independent components. First, we introduce a lightweight Fine-grained Consistency LoRA (FCLoRA), which performs a one-time refinement of a pretrained subject-to-image model within only a few seconds to improve fine-grained subject consistency. 
Second, we revisit the design of Subject-to-Image LoRA and propose a Visual-only LoRA strategy. Specifically, we remove LoRA from the text stream of the Double-Stream Blocks while retaining it in the image stream and the subsequent Single-Stream Blocks.

\subsection{Fine-grained Consistency LoRA}

\subsubsection{Motivation.}

Existing subject-to-image models reconstruct a target image from one or more reference images. Although the reference and target images share the same identity, they usually differ in pose, viewpoint, lighting, and many other appearance factors. Consequently, the training objective primarily encourages semantic identity alignment instead of establishing precise fine-grained correspondence between the reference and target images.
As a result, current subject-to-image models generally preserve the overall identity of a subject while often failing to faithfully reproduce subtle subject-specific details. These missing details accumulate into noticeable identity inconsistency during generation.

Instead of improving supervision through additional large-scale synthesized data, we ask whether a pretrained subject-to-image model can recover these missing fine-grained details through an extremely lightweight refinement. Our key observation is that if the conditioning image and the reconstruction target are exactly the same image, all appearance variations disappear and the optimization objective becomes a pure fine-grained reconstruction problem. Since semantic identity alignment has already been learned during subject-to-image training, the refinement only needs to strengthen the model's ability to preserve fine-grained visual correspondence, making a lightweight refinement completed within only a few seconds sufficient.
Motivated by this observation, we introduce a Fine-grained Consistency LoRA (FCLoRA), which performs a second-stage lightweight refinement on top of an existing Subject-to-Image LoRA.

\subsubsection{Fine-grained Consistency Refinement.}

Given a pretrained subject-to-image model with parameters
$\theta=\theta_{\mathrm{base}}+\Delta\theta_{\mathrm{S2I}}$,
where $\theta_{\mathrm{base}}$ denotes the frozen text-to-image backbone and
$\Delta\theta_{\mathrm{S2I}}$ represents the pretrained Subject-to-Image LoRA, we further attach a lightweight Fine-grained Consistency LoRA
$\Delta\theta_{\mathrm{FC}}$,
whose rank is significantly smaller than that of the original Subject-to-Image LoRA.

Unlike conventional subject-to-image training, where the conditioning image and reconstruction target depict the same subject but differ in appearance, our refinement adopts an exact self-reconstruction objective by constructing
$I_{\mathrm{ref}} = I_{\mathrm{target}}$,
where the conditioning image and the reconstruction target are exactly identical, including both pixel values and image resolution.
The refinement objective is formulated as

\begin{equation}
\mathcal{L}_{FC}
=
\mathcal{L}_{FM}
\left(
f_{\theta_{\mathrm{base}}
+\Delta\theta_{\mathrm{S2I}}
+\Delta\theta_{\mathrm{FC}}}
(I_{\mathrm{ref}}),
I_{\mathrm{target}}
\right),
\end{equation}
where $\mathcal{L}_{FM}$ is the original Flow Matching objective.

Compared with conventional subject-to-image training, the proposed objective completely removes appearance discrepancies between the conditioning image and the reconstruction target. Consequently, the refinement no longer needs to model pose variation, viewpoint transformation, illumination changes, or other semantic correspondence, allowing the lightweight FCLoRA to focus exclusively on strengthening fine-grained visual correspondence.

More importantly, we find that this exact correspondence is essential for effective refinement. Any mismatch between the conditioning image and the reconstruction target, including using another image of the same subject or changing the image resolution, causes the refinement to fail to produce noticeable improvement. Owing to the greatly simplified objective, the proposed FCLoRA completes the refinement within only a few seconds while consistently improving fine-grained subject consistency.

\subsubsection{Generalization.}

Although FCLoRA is optimized using an exact self-reconstruction proxy task, the refinement does not encode the identity or appearance of the proxy image itself. Instead, the proxy image provides a stable optimization objective that exposes the fine-grained consistency errors already present in the pretrained subject-to-image model. By correcting these errors through a lightweight model-level refinement, FCLoRA improves how the model preserves subtle visual details during subject-conditioned generation, rather than learning image-specific features. Consequently, the refined model consistently improves subject consistency for unseen identities and prompts without requiring subject-specific optimization.

\subsection{Visual-only Subject-to-Image LoRA}

Existing subject-to-image models built upon the FLUX~\cite{flux2024} backbone typically apply LoRA to both the text and image streams of the Double-Stream Blocks, as well as the subsequent Single-Stream Blocks. However, subject-specific knowledge is primarily encoded in visual appearance, while the textual prompt mainly specifies the desired scene or attributes. This motivates us to revisit whether adapting the text stream is actually necessary for subject-to-image learning.
Formally, the original Subject-to-Image LoRA is
\begin{equation}
\Delta\theta_{\mathrm{S2I}}
=
\Delta\theta_{\mathrm{D}}^{\mathrm{text}}
+
\Delta\theta_{\mathrm{D}}^{\mathrm{image}}
+
\Delta\theta_{\mathrm{S}},
\end{equation}
where $\Delta\theta_{\mathrm{D}}^{\mathrm{text}}$ and
$\Delta\theta_{\mathrm{D}}^{\mathrm{image}}$ denote the LoRA parameters inserted into the text and image streams of the Double-Stream Blocks, respectively, while $\Delta\theta_{\mathrm{S}}$ denotes the LoRA parameters in the subsequent Single-Stream Blocks.

Instead of adapting both streams, we remove LoRA only from the text stream while retaining all visual adaptation modules, resulting in
\begin{equation}
\Delta\theta_{\mathrm{S2I}}^{\mathrm{VO}}
=
\Delta\theta_{\mathrm{D}}^{\mathrm{image}}
+
\Delta\theta_{\mathrm{S}}.
\end{equation}
All remaining training settings are kept unchanged. This design preserves the pretrained text representations while allowing subject-specific adaptation to be learned through the visual stream and the subsequent shared Single-Stream Blocks.

%% file: sec/4_exp.tex
\begin{table}[t]
    \centering
    \tabcolsep=0.14cm
    \begin{tabular}{lcccc}
    \toprule
    Method & ID$_{\text{single}}$ $\uparrow$ & IP$_{\text{single}}$ $\uparrow$ & ID$_{\text{multi}}$ $\uparrow$ & IP$_{\text{multi}}$ $\uparrow$ \\
    \midrule
    MIP-Adapter & 39.59 & 71.97 & 24.58 & 57.00 \\
    OmniGen & 76.51 & 78.46 & 55.53 & 62.32 \\
    OmniGen2 & 62.41 & 74.08 & 40.81 & 67.15 \\
    DreamO & 75.48 & 70.84 & 50.24 & 64.63 \\
    XVerse & 79.48 & 76.86 & \textbf{66.59} & 71.48 \\
    \midrule
    UNO & 59.24 & 80.75 & 37.62 & 68.30 \\
    UNO+CopyCat & 69.31 & \textbf{84.95} & 46.69 & \textbf{72.61} \\
    \cmidrule(l){1-5}
    UMO & 75.16 & 78.88 & 58.69 & 64.81 \\
    UMO+CopyCat & \textbf{80.25} & 82.79 & 63.29 & 69.29 \\
    \cmidrule(l){1-5}
    USO & 60.98 & 79.44 & 43.04 & 62.74 \\
    USO+CopyCat & 63.35 & 80.12 & 45.08 & 63.76 \\
    \bottomrule
    \end{tabular}
    \caption{\textbf{Comparison on XVerseBench.}
CopyCat consistently improves subject consistency under both single-subject and multi-subject settings.}
    \label{tab:xverse}
\end{table}

\begin{table}[t]
    \centering
    \tabcolsep=0.4cm
    \begin{tabular}{lccc}
    \toprule
    Method & I$_{\text{DINO}}$ $\uparrow$ & I$_{\text{CLIP}}$ $\uparrow$ & T$_{\text{CLIP}}$ $\uparrow$ \\
    \midrule
    RealCustom++ & 0.702 & 0.794 & 0.318 \\
    OmniGen & 0.693 & 0.801 & 0.315 \\
    OminiControl & 0.684 & 0.799 & 0.312 \\
    \midrule
    UNO & 0.754 & 0.837 & 0.302 \\
    UNO+CopyCat & \textbf{0.800} & \textbf{0.851} & 0.300 \\
    \cmidrule(l){1-4}
    UMO & 0.754 & 0.835 & 0.298 \\
    UMO+CopyCat & 0.796 & \textbf{0.851} & 0.296 \\
    \cmidrule(l){1-4}
    USO & 0.750 & 0.828 & \textbf{0.320} \\
    USO+CopyCat & 0.766 & 0.832 & 0.319 \\
    \bottomrule
    \end{tabular}
    \caption{\textbf{Single-subject comparison on DreamBench.}
CopyCat consistently improves identity preservation while maintaining comparable text alignment.}
    \label{tab:db_single}
\end{table}

\begin{table}[t]
    \centering
    \tabcolsep=0.36cm
    \begin{tabular}{lccc}
    \toprule
    Method & I$_{\text{DINO}}$ $\uparrow$ & I$_{\text{CLIP}}$ $\uparrow$ & T$_{\text{CLIP}}$ $\uparrow$ \\
    \midrule
    Subject Diffusion & 0.506 & 0.696 & 0.310 \\
    MIP-Adapter & 0.482 & 0.726 & 0.311 \\
    MS-Diffusion & 0.525 & 0.726 & 0.319 \\
    OmniGen & 0.542 & 0.733 & 0.322 \\
    \midrule
    UNO & 0.531 & 0.731 & 0.320 \\
    UNO+CopyCat & \textbf{0.556} & \textbf{0.737} & 0.320 \\
    \cmidrule(l){1-4}
    UMO & 0.527 & 0.727 & 0.316 \\
    UMO+CopyCat & 0.552 & 0.731 & 0.314 \\
    \cmidrule(l){1-4}
    USO & 0.496 & 0.719 & \textbf{0.340} \\
    USO+CopyCat & 0.502 & 0.719 & 0.338 \\
    \bottomrule
    \end{tabular}
    \caption{\textbf{Multi-subject comparison on DreamBench.}
CopyCat consistently improves multi-subject identity preservation while preserving text alignment.}
    \label{tab:db_multi}
\end{table}

\begin{figure*}[t]
    \centering
    \includegraphics[width=\linewidth]{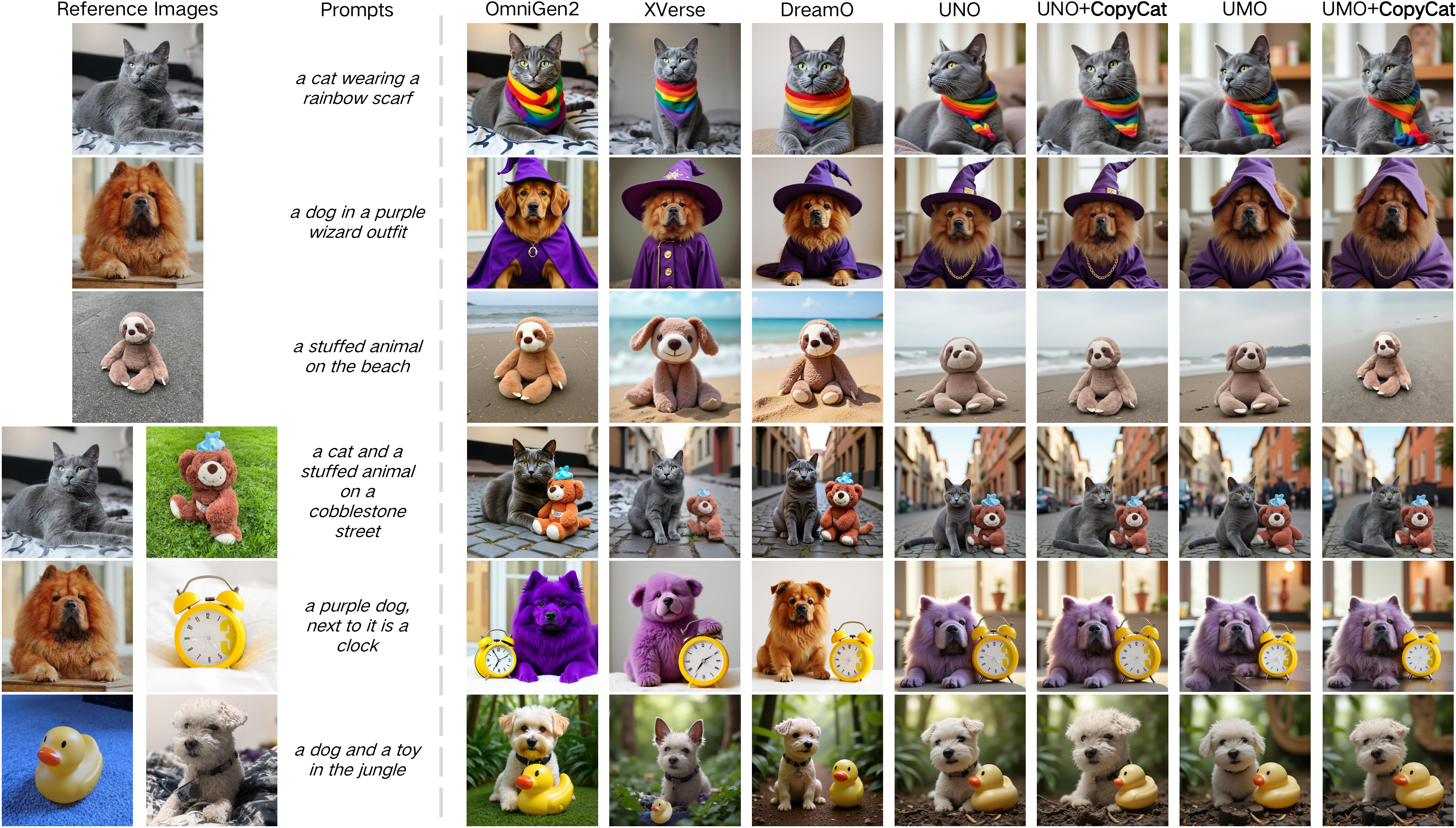}
    \caption{\textbf{Qualitative comparisons on single-subject and multi-subject personalization.} Compared with representative subject-to-image personalization models, CopyCat better preserves fine-grained subject appearance while faithfully following the target prompt. Zoom in for better visualization.}
    \label{fig:comparison}
\end{figure*}

\begin{figure}[t]
    \centering
    \includegraphics[width=\linewidth]{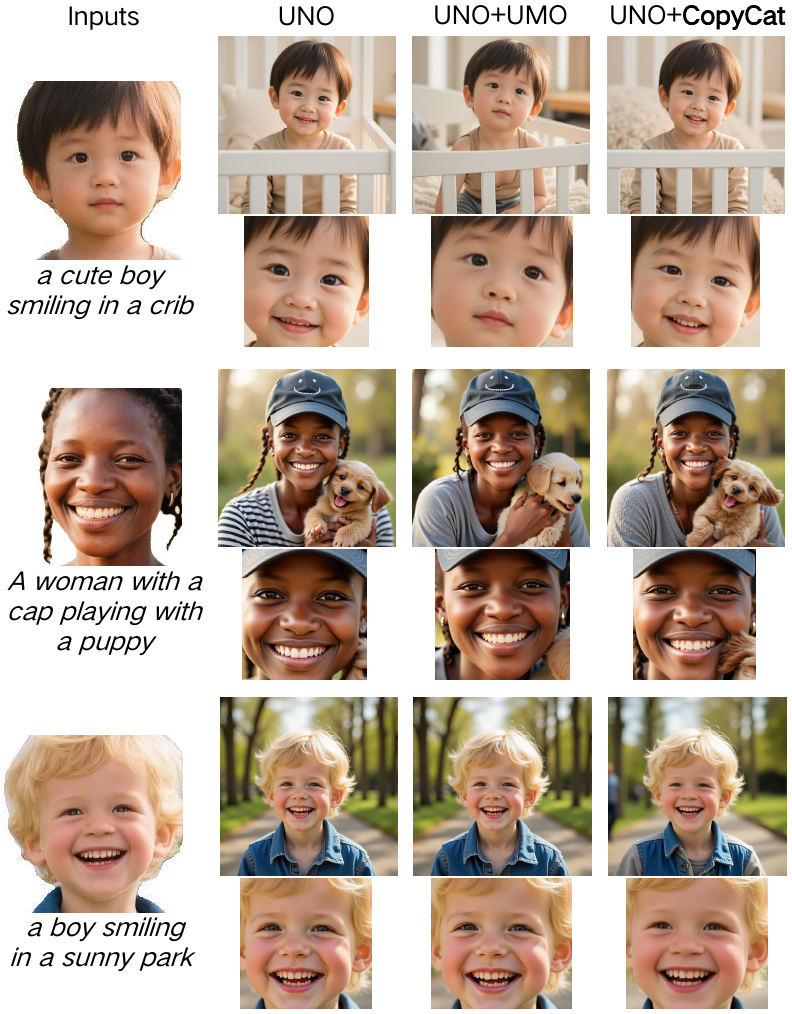}
    \caption{\textbf{Comparison of Human Identity Consistency Improvement.}
CopyCat substantially improves facial identity consistency and achieves performance comparable to UMO without using any face-specific training data.}
    \label{fig:face_consistency}
\end{figure}

\begin{figure}[t]
    \centering
    \includegraphics[width=\linewidth]{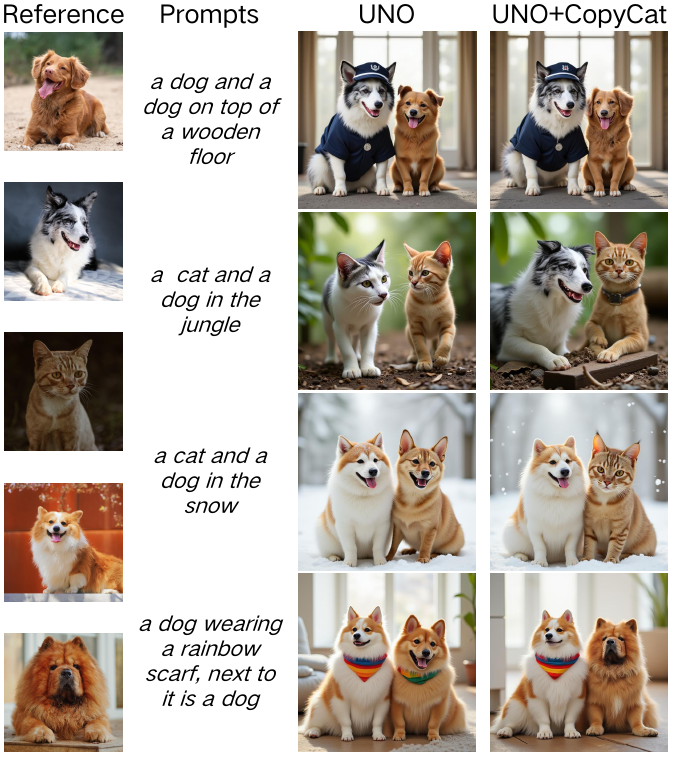}
    \caption{\textbf{Alleviating identity confusion between similar-category subjects.}
CopyCat effectively mitigates identity confusion when composing visually similar subjects by preserving the fine-grained characteristics of each subject.}
    \label{fig:id_confusion}
\end{figure}


\begin{figure*}[t]
    \centering
    \includegraphics[width=\linewidth]{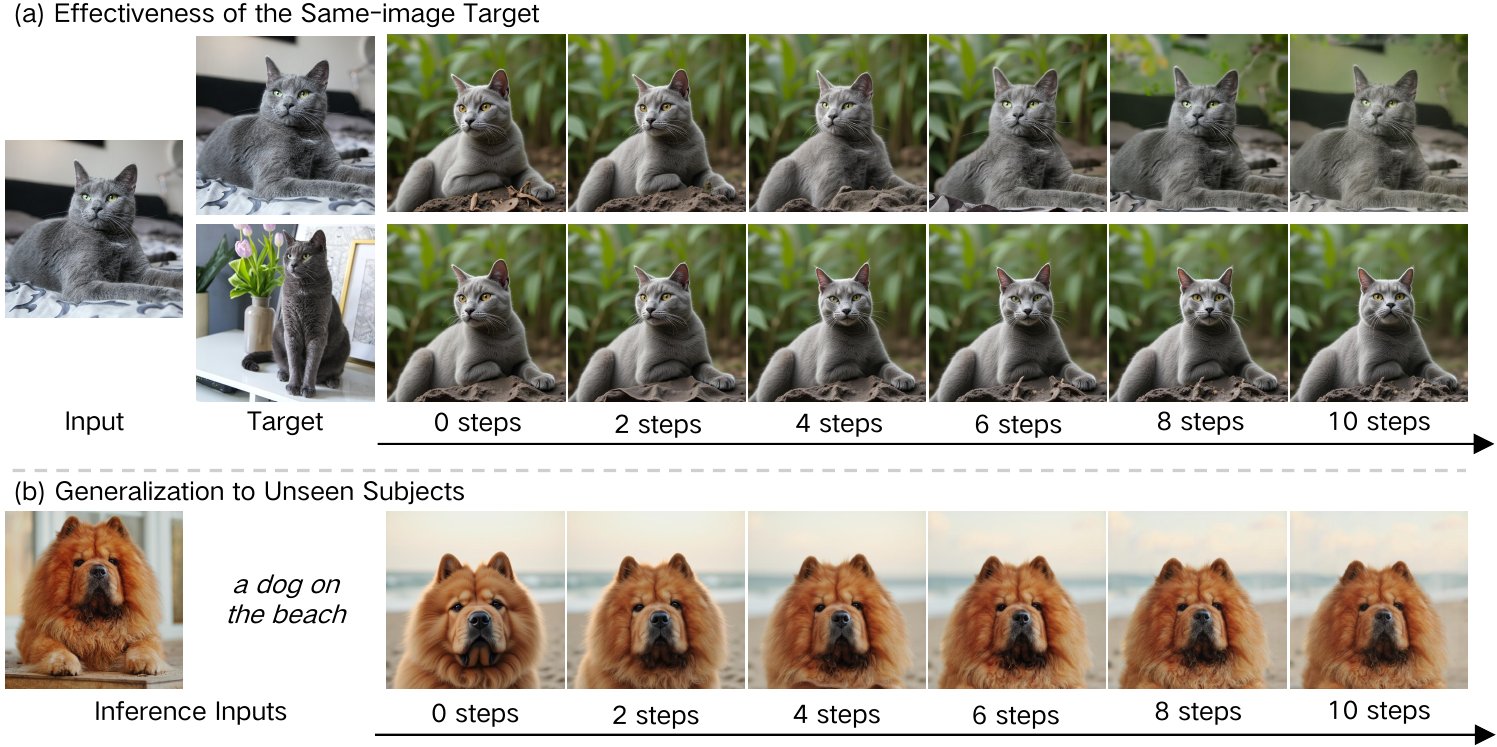}
\caption{\textbf{Importance of the same-image refinement target.} 
(a) Comparison between the proposed same-image target and using a different image of the same subject as the refinement target. The proposed same-image target rapidly improves fine-grained subject consistency within only a few optimization steps, whereas using another image of the same subject fails to produce noticeable improvement. Excessive optimization may gradually overfit image-specific appearance.
(b) Generalization to unseen subjects. Models obtained after different optimization steps are evaluated on an unseen subject. The consistent improvement in identity preservation demonstrates that the learned fine-grained consistency capability generalizes beyond the proxy image.
}
\label{fig:refinement_steps}
\end{figure*}

\begin{table}[t]
    \centering
    \begin{tabular}{lcc|cc}
    \toprule
    & \multicolumn{2}{c|}{XVerseBench} & \multicolumn{2}{c}{DreamBench} \\
    \cmidrule(lr){2-3}\cmidrule(l){4-5}
    Model & ID$_{\text{single}}$ $\uparrow$ & IP$_{\text{single}}$ $\uparrow$ & I$_{\text{DINO}}$ $\uparrow$ & I$_{\text{CLIP}}$ $\uparrow$ \\
    \midrule
    Full LoRA & 29.13 & 70.17 & 0.6562 & 0.7908 \\
    Visual-only & \textbf{38.66} & \textbf{73.33} & \textbf{0.7033} & \textbf{0.8157} \\
    \bottomrule
    \end{tabular}
    \caption{\textbf{Effect of the Visual-only LoRA strategy.}
Both models are retrained on limited training data and are therefore not directly comparable to the official UNO checkpoint.}
    \label{tab:visual_only}
\end{table}

\section{Experiments}

\subsection{Experimental Setup}

\subsubsection{Benchmarks.}

We evaluate CopyCat on both DreamBench~\cite{ruiz2023dreambooth} and XVerseBench~\cite{chen2026xverse}. DreamBench is the most widely adopted benchmark for subject personalization. Following the standard protocol, we report I$_{\rm DINO}$ and I$_{\rm CLIP}$ to measure subject consistency, and T$_{\rm CLIP}$ to measure prompt consistency. We further evaluate on XVerseBench, a recently proposed benchmark designed for modern in-context subject generation models. XVerseBench covers both single-subject and multi-subject settings across diverse subject categories. Following its official protocol, we report IP$_{\rm single}$, IP$_{\rm multi}$, ID$_{\rm single}$, and ID$_{\rm multi}$, where IP measures general subject consistency, while ID specifically evaluates human identity consistency.


\subsubsection{Experimental Protocol.}

All CopyCat results reported in the paper are obtained using a single refined model. FCLoRA is optimized only once using a single proxy image and is directly applied to all unseen reference subjects without subject-specific refinement. We also experiment with different proxy images and observe nearly identical performance. Since only a lightweight rank-4 LoRA is optimized for four optimization steps, the entire refinement is completed in approximately 3 seconds on a single GPU. The Visual-only Subject-to-Image LoRA is trained for 20K optimization steps using four GPUs.
We compare with publicly available models. FocusDPO and PSR are not included since their model checkpoints are not publicly available.

\subsection{Comparisons}

\subsubsection{Subject-driven Generation.}

CopyCat is designed as a lightweight refinement strategy for existing subject-to-image models. We evaluate it on three representative personalization models, including UNO, UMO, and USO. Figure~\ref{fig:comparison} and Tables~\ref{tab:xverse}, \ref{tab:db_single}, and \ref{tab:db_multi} present the qualitative and quantitative comparisons on both single-subject and multi-subject generation.
Compared with their corresponding backbone models, CopyCat consistently preserves finer appearance details while producing more faithful multi-subject compositions. Consistent improvements are observed across all three backbones on both XVerseBench and DreamBench, demonstrating the effectiveness of the proposed Fine-grained Consistency LoRA across different subject-to-image models.

Beyond consistently improving existing models, the refined models also achieve superior performance compared with representative state-of-the-art subject-to-image methods. In particular, CopyCat achieves stronger subject consistency while maintaining comparable text alignment, indicating that the proposed refinement strategy effectively enhances existing personalization models without sacrificing generation quality.

\subsubsection{Human Face Personalization.}

Although CopyCat is trained without any face-specific supervision, we further evaluate its performance on human face personalization. As reported in Table~\ref{tab:xverse}, CopyCat consistently improves the identity similarity metric (ID) over baselines. Figure~\ref{fig:face_consistency} further shows that the improved identity consistency is reflected in more faithful facial characteristics, achieving results comparable to UMO despite the absence of face-specific enhancement. These results demonstrate that the proposed refinement strategy naturally generalizes to human face personalization.

\subsection{Further Analysis}

\subsubsection{Identity Confusion Mitigation.} 
Identity confusion is a common failure mode in multi-subject personalization, particularly for visually similar subjects. As shown in Figure~\ref{fig:id_confusion}, UNO often mixes the identities of different subjects, leading to ambiguous appearances. CopyCat effectively separates individual identities and preserves their distinctive visual characteristics, substantially reducing identity confusion.

\subsubsection{Effect of the Refinement Target.}
Figure~\ref{fig:refinement_steps} analyzes the refinement target and duration. As shown in Figure~\ref{fig:refinement_steps}(a), using another image of the same subject as the reconstruction target brings little benefit, whereas the proposed identical-image target noticeably improves subject consistency within only a few refinement steps. Longer refinement gradually overfits to image-specific details, suggesting that only a lightweight refinement is necessary. Figure~\ref{fig:refinement_steps}(b) further demonstrates that the learned consistency capability generalizes to unseen subjects rather than memorizing the proxy image.

\subsubsection{Effect of Visual-only LoRA.}
Table~\ref{tab:visual_only} compares the conventional Full LoRA with the proposed Visual-only LoRA. Since both models are retrained on our available training data, their absolute performance is not directly comparable to the official UNO checkpoint. Nevertheless, Visual-only LoRA consistently improves subject consistency on both benchmarks. Additional qualitative comparisons are provided in the supplementary material. These results indicate that adapting the text stream is unnecessary for subject-specific personalization, as subject-specific visual characteristics can be effectively learned through the image branch together with the subsequent shared Single-Stream Blocks.

%% file: sec/5_conclusion.tex
\section{Conclusion}

In this paper, we presented CopyCat, a lightweight refinement framework for improving fine-grained subject consistency in subject-to-image personalization models. We showed that a simple proxy task based on identical image reconstruction is sufficient to learn a transferable fine-grained consistency capability. This capability consistently enhances identity preservation across multiple personalization models after only a few seconds of refinement. We further demonstrated that a simple Visual-only LoRA design consistently improves subject consistency by avoiding unnecessary adaptation of the text stream. Extensive experiments on XVerseBench and DreamBench validate the effectiveness and generality of both designs. We hope the proposed refinement strategy provides a simple and practical direction for improving future subject-to-image personalization models.